\documentclass[sigconf]{acmart}
\AtBeginDocument{%
  }

\setcopyright{acmlicensed}
\copyrightyear{2025}
\acmYear{2025}
\acmDOI{10.1145/XXXXXX.XXXXXX}
\acmConference[MM '25] {Proceedings of the 33rd ACM International Conference on Multimedia}{October 27--31, 2025}{Dublin, Ireland.}
\acmBooktitle{Proceedings of the 33rd ACM International Conference on Multimedia (MM '25), October 27--31, 2025, Dublin, Ireland}
\acmISBN{979-8-4007-2035-2/2025/10}
\usepackage{xcolor} 
\usepackage{pifont}
\usepackage{colortbl} 
\usepackage{multirow}
\usepackage{graphicx}  
\usepackage{makecell}

\begin{document}

\title{\includegraphics[height=0.8em]{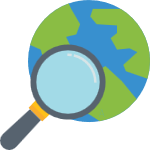}GeoMag: A Vision-Language Model for Pixel-level Fine-Grained Remote Sensing Image Parsing}


\author{Xianzhi Ma}
\orcid{0009-0007-4909-0894}
\affiliation{%
  \institution{Institute of Space Earth Science, School of Frontier Sciences, Nanjing University}
  \city{Suzhou}
  \country{China}}
\email{xianzhima@smail.nju.edu.cn}

\author{Jianhui Li}
\orcid{0009-0001-6253-9808}
\affiliation{%
  \institution{Institute of Space Earth Science, School of Frontier Sciences, Nanjing University}
  \city{Suzhou}
  \country{China}}
\email{lijh@nju.edu.cn}
\authornote{Corresponding author.}

\author{Changhua Pei}  
\orcid{0000-0001-9288-4787}
\affiliation{%
  \institution{Computer Network Information Center, Chinese Academy of Sciences}
  \city{Beijing}
  \country{China}}
\email{chpei@cnic.cn}

\author{Hao Liu}    
\orcid{0000-0002-0411-5396}
\affiliation{%
  \institution{Institute of Space Earth Science, School of Frontier Sciences, Nanjing University}
  \city{Suzhou}
  \country{China}}
\email{haoliu@nju.edu.cn}


\begin{abstract}
The application of Vision-Language Models (VLMs) in remote sensing (RS) image understanding has achieved notable progress, demonstrating the basic ability to recognize and describe geographical entities. However, existing RS-VLMs are mostly limited to image-level and region-level tasks, lacking the capability to handle pixel-level tasks and performing poorly in small-object recognition scenarios. Moreover, RS-VLMs consume significant computational resources when processing high-resolution RS images, further restricting their practical applicability. In this context, we propose GeoMag (Geographical Magnifier), an end-to-end general-purpose large model framework for RS. GeoMag dynamically focuses the attention scope based on prompt semantics to effectively perform remote sensing image parsing across multiple levels of granularity. This method introduces Task-driven Multi-granularity Resolution Adjustment and Prompt-guided Semantic-aware Cropping, which adaptively reduce the spatial resolution of task-irrelevant regions while enhancing the visual representation of task-relevant areas. This approach improves the model's perception of critical target regions, suppresses background redundancy, and reduces the computational cost of interpreting high-resolution RS imagery. Extensive comparative experiments on 10 benchmarks demonstrate that GeoMag not only excels in handling pixel-level tasks but also maintains competitive performance across tasks of other granularities compared to existing RS-VLMs.
\end{abstract}

\ccsdesc[500]{Information systems~Image search}
\ccsdesc[500]{Information systems~Multimedia and multimodal retrieval}

\keywords{VLMs; Remote Sensing Image; Pixel-level; Attention Mechanism.}


\maketitle

\section{Introduction}

\begin{figure}[h]
  \includegraphics[width=1\linewidth]{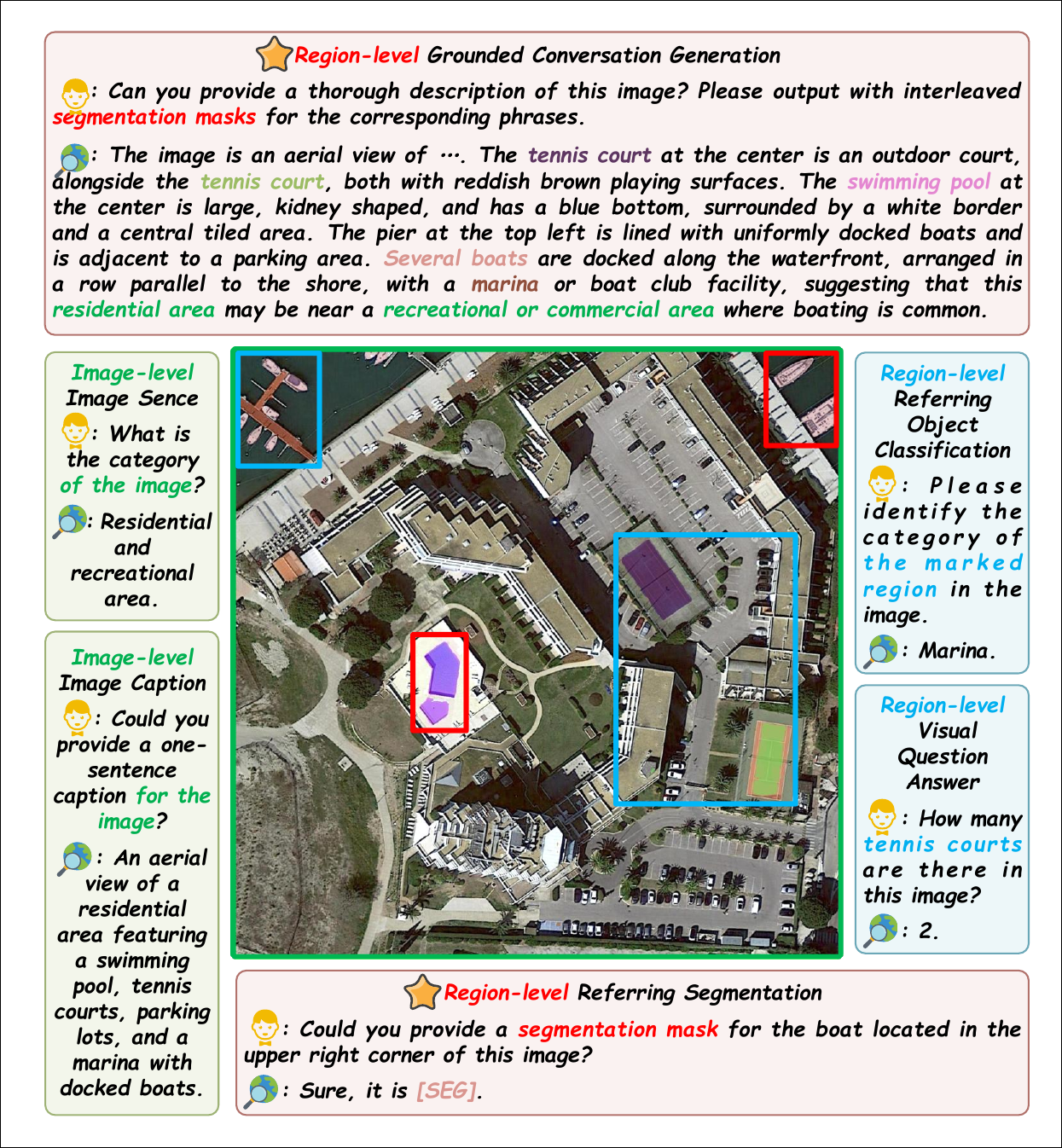}
  \caption{Overview of GeoMag's capabilities. In addition to understanding both global and local image information for image-level tasks (green box) and region-level tasks (blue box), GeoMag extends the functionality of existing RS-VLMs by enabling pixel-level task processing—an ability typically absent in current models. As illustrated by the red box, GeoMag can automatically generate detailed descriptions of the image and produce segmentation masks for key geographical entities mentioned in the text. It can also perform user-directed segmentation based on specified prompts.}
  \label{fig:1}
\end{figure}

Remote sensing (RS) imagery, as a vital source of Earth observation data, serves scientific research, decision-making support, and the monitoring and evaluation of Sustainable Development Goals (SDGs) \citep{sdg}. Therefore, effectively understanding the rich semantic content embedded in RS images becomes a fundamental step toward unlocking the value of RS data and promoting the intelligent development of earth science research \citep{sscnet, stripunet}.

RS image parsing tasks can be categorized based on the level of granularity into image-level (coarse-grained), region-level (mid-grained), and pixel-level (fine-grained) tasks \citep{survey, GeoPixel, GeoPix, EarthMarker}. These correspond respectively to the interpretation of entire images, the identification and localization of specific regions or objects, and the precise annotation of individual pixels. With the advancement of Vision-Language Models (VLMs), RS image interpretation is undergoing a paradigm shift toward multimodal integration \citep{mmrh1, mmrh2, mmrh3, mmrh4, mmrh5, mmrh6}. Within this context, research on pixel-level tasks (e.g., semantic segmentation, change detection) and small-object monitoring (e.g., vehicle and ship recognition) holds significant importance for building collaborative RS-language intelligent systems and achieving natural language-driven automated image analysis \citep{od, rseg1, rseg2}.

Current RS-VLMs have already achieved noticeable progress in coarse-grained and mid-grained RS image parsing tasks \citep{EarthGPT, RemoteCLIP, GeoRSCLIP}. These tasks typically include: (a) image-level tasks such as scene classification, image captioning, and image retrieval; and (b) region-level tasks such as visual grounding, object classification, and object counting. These general-purpose RS-VLMs even outperform task-specific models in many cases, suggesting that VLMs hold great potential for extracting the rich information embedded in RS imagery \citep{SkySense, GRAFT, SpectralGPT}. However, most publicly available RS-VLMs lack the ability to handle pixel-level tasks. These models fail to provide precise analysis and detailed processing of RS imagery, which limits their applicability in high-precision scenarios such as land cover extraction and pixel-level change monitoring \citep{SkyEyeGPT, VHM, LHRS}.

In addition to the lack of pixel-level processing capabilities, current RS-VLMs also face challenges in parsing small objects within RS images \citep{EarthMarker, GeoChat}. Beyond the inherent difficulties posed by small objects—such as limited size, sparse features, and susceptibility to interference from complex backgrounds—the feature extraction strategies adopted by existing RS-VLMs generally lack regional focus. These models fail to highlight the deep semantic association between the task prompt and the image content, which is a key factor limiting their ability to recognize small targets \citep{EarthGPT, RemoteCLIP, GeoRSCLIP, SkySense}. Furthermore, the emphasis on generalized, holistic feature extraction strategies often involves directly encoding the entire original image \citep{GRAFT, SkyEyeGPT, VHM, LHRS}, resulting in substantial computational overhead and further constraining the efficient application of RS-VLMs in fine-grained remote sensing scenarios.

To address these issues, we propose GeoMag (Geographical Magnifier), an end-to-end general-purpose large model framework for remote sensing. As illustrated in Figure~\ref{fig:1}, GeoMag performs pixel-level tasks by identifying key geographical entities within an image and generating corresponding descriptions and segmentation masks, demonstrating strong segmentation performance for small objects such as ships.

The contributions of this study are summarized as follows:

1. We propose GeoMag, an RS-VLM capable of performing remote sensing image parsing across multiple granularities, including image level, region level, and pixel level, thereby addressing the current inability of foundational RS models to process pixel-level tasks.

2. We introduce Task-driven Multi-granularity Resolution Adjustment and Prompt-guided Semantic-aware Cropping, which emphasize task-relevant regions within the image, enhance the recognition of small objects, and reduce computational cost.

3. We conducted extensive experiments across 10 benchmarks, and results show that GeoMag not only possesses strong pixel-level processing capabilities but also outperforms existing RS-VLMs on tasks of other granularities.

\section{Related Works}
\textbf{Pixel-level Parsing.} In computer vision \citep{buchong1, buchong2, buchong3}, pixel-level RS image parsing tasks mainly include Referring Segmentation and Open Vocabulary Segmentation \citep{survey, rseg1, rseg2}. To enable large language models (LLMs) to perform pixel-level tasks, PixelLM \citep{PixelLM} and LISA \citep{LISA} introduced a [SEG] token into the vocabulary of the LLM as a conditional input, allowing the decoder to generate segmentation masks from textual descriptions. GSVA \citep{GSVA} further introduced a [REJ] token to explicitly exclude irrelevant or unwanted targets, enabling the model to follow user intent more precisely. GLaMM \citep{GLaMM} proposed a novel pixel-level task called Grounded Conversation Generation (GCG), aiming to identify and localize specific regions or objects in images through human-machine dialogue, thereby opening a promising new direction for pixel-level task research. However, these studies were conducted on natural images, which differ significantly from RS images in terms of semantic content and visual characteristics. Therefore, directly transferring such methods often fails to yield satisfactory results \citep{qianyibuhao}, and specialized optimization and adaptation for RS imagery are necessary.

\textbf{RS-VLM.} With the integration of multimodal techniques into the RS domain \citep{AMFMN, RemoteCLIP, GeoRSCLIP, GRAFT, UrbanCross}, existing RS-VLMs are able to perform both image-level and region-level tasks. As early RS-VLMs, RSGPT \citep{RSGPT} and RS-CapRet \citep{RS-CapRet} successfully combined LLMs to complete image-level tasks such as image captioning and image retrieval. LHRS-Bot \citep{LHRS}, GeoChat \citep{GeoChat}, and EarthMarker \citep{EarthMarker} extended this foundation to achieve region-level understanding. Further, SkyEyeGPT \citep{SkyEyeGPT}, SkySense \citep{SkySense}, and EarthGPT \citep{EarthGPT} refined region-level tasks by enabling object localization and detection. However, current RS-VLMs typically place the LLM at the end of the processing pipeline \citep{EarthGPT, SkyEyeGPT, VHM, LHRS, GeoChat, EarthMarker}, which limits output to text and prevents pixel-level task execution. Therefore, as in models like PixelLM \citep{PixelLM}, it is necessary to append a pixel decoder—such as those in the SAM family \citep{sam, sam2} or Mask2Former \citep{Mask2Former}—after the LLM to enable segmentation.

\begin{figure*}[t]
  \includegraphics[width=1\linewidth]{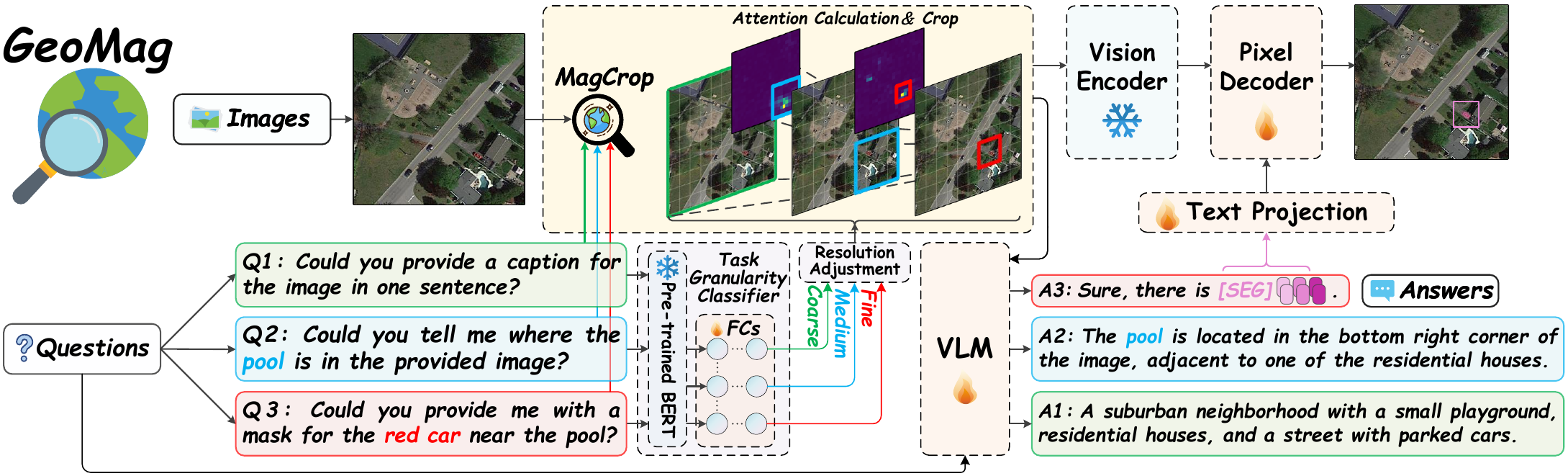}
  \caption {GeoMag overview. The framework classifies the input query by task granularity, adjusts the resolution of the input image based on the classification result, and performs attention-based querying and cropping between the query and the image. This process precisely highlights important regions in the image and improves the model's efficiency in handling high-resolution inputs. The refined query and image are then fed into the VLM to generate more accurate responses. For pixel-level tasks, the framework further feeds the adjusted image into a vision encoder. The output of the vision encoder, together with the [SEG] token generated by the VLM, is passed into a pixel decoder to produce the segmentation mask.}
  \label{fig:2}
\end{figure*}

\section{GeoMag}

As shown in Figure~\ref{fig:2}, the structure and workflow of GeoMag primarily involve two novel components: Task-driven Multi-granularity Resolution Adjustment and Prompt-guided Semantic-aware Cropping. These modules help emphasize task-relevant regions in remote sensing images while suppressing irrelevant background noise, offering a precise and efficient solution for RS image parsing.

\subsection{GeoMag Architecture for Enabling Pixel-level Parsing}

Specifically, GeoMag employs a frozen SAM2 \citep{sam2} as the vision encoder, leveraging its powerful general-purpose image feature extraction capabilities to generate multi-scale segmentation features $f_{img}^l$(l=1,2,...,L), which assist the downstream pixel decoder in achieving accurate segmentation. In addition to the multi-scale features extracted from SAM2, the pixel decoder also receives $N$ segmentation tokens $h_{seg}^n$(n=1,2,...,N) output by the VLM. These tokens contain contextual information about the segmentation targets. GeoMag further designs a Text Projection layer, consisting of a two-layer MLP, to project the 4096-dimensional context information output by the VLM into a 256-dimensional space suitable for processing by the pixel decoder. This ensures the model can accurately interpret and generate segmentation masks even in complex scenes. The pixel decoder integrates segmentation features from different scales and generates full-scale segmentation masks through weighted averaging as the final output. This mask generation process is formally expressed in Equation \ref{eq:1}:
\begin{equation}
  \label{eq:1}
\text { Mask }=\sum_{\mathrm{l}=1}^{\mathrm{L}} \omega_l \cdot \text { Pixel Decoder }\left(\sum_{\mathrm{n}=1}^{\mathrm{N}} \beta_{\mathrm{n}} \cdot \mathrm{~h}_{\text {seg }}^{\mathrm{n}}, \mathrm{f}_{\text {img }}^{\mathrm{l}}\right)
\end{equation}

Here, $\omega_l$(l=1,2,...,L) and $\beta_{\mathrm{n}}$(n=1,2,...,N) represent the weighting factors corresponding to the segmentation features $f_{img}^l$ and the segmentation tokens $h_{seg}^n$, respectively, and are used to control their relative importance. In addition, both the Text Projection layer and the Pixel Decoder are trainable, enabling GeoMag to better adapt to the specific demands of remote sensing tasks.

\subsection{MagCrop}

\begin{figure*}[t]
  \includegraphics[width=1\linewidth]{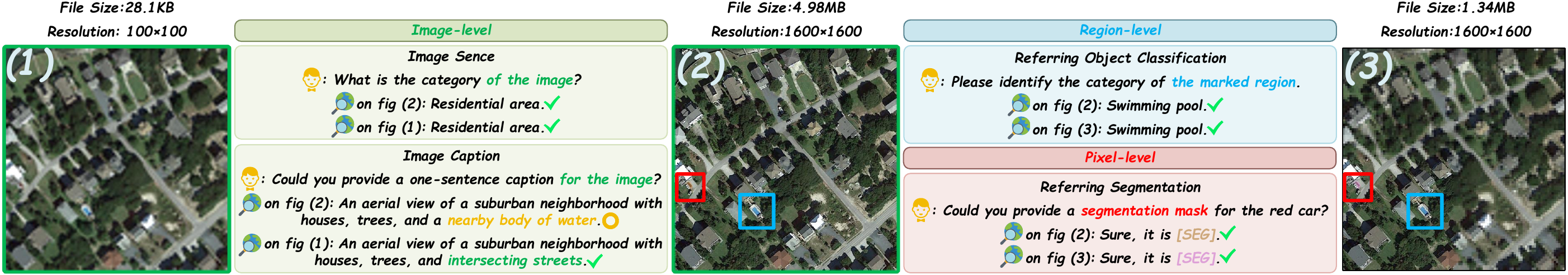}
  \caption {The impact of different image preprocessing methods on tasks of varying granularity.
In the figure, (1) represents an image with reduced resolution, (2) is the original remote sensing image, and (3) is an image where the key regions remain unchanged while task-irrelevant areas are compressed. As shown, using image (1) for image-level tasks and using image (3) for region-level and pixel-level tasks yields results comparable to—or even better than—those using the original image (2). The advantage becomes evident in the Image Caption task, where the blurred version of the image in (1) leads GeoMag to ignore the detail of "water body" and instead focus on the overall structure of the image, resulting in a higher-level description like "intersecting streets" that aligns more appropriately with the granularity of the task.}
  \label{fig:3}
\end{figure*}

For a general-purpose remote sensing foundation model designed to support multi-granularity tasks, the strategies for processing images and queries should be diverse. As shown in Figure~\ref{fig:3}, the file size of the original image (2) is approximately 181 and 3.7 times larger than that of image (1) and image (3), respectively, and the number of visual tokens input to the VLM is 256 times that of image (1). However, by examining the outputs of GeoMag across various task granularities, it is observed that the processed images not only achieve comparable results to the original inputs, but even yield more appropriate outcomes for the Image Caption task, aligning better with the intended granularity.

To more effectively address challenges in remote sensing image parsing—such as varying granularity, sparse targets, and uneven spatial distributions—this work proposes an image preprocessing strategy named MagCrop. MagCrop reduces resource consumption (e.g., memory, GPU usage, and time) and enhances model response accuracy through the following principles: (1) For tasks emphasizing global features, it appropriately reduces image resolution. (2) For tasks with explicit spatial references (e.g., specific locations or objects), it preserves the resolution of task-relevant regions while compressing irrelevant areas. MagCrop integrates two key mechanisms: Task-driven Multi-granularity Resolution Adjustment and Prompt-guided Semantic-aware Cropping. These mechanisms help maintain the resolution of task-relevant regions while effectively reducing computational redundancy and emphasizing critical semantic content.

\subsubsection{Task-driven Multi-granularity Resolution Adjustment}
Existing RS-VLMs typically input raw images directly into the VLM or adopt a fixed preprocessing strategy \citep{EarthGPT, RemoteCLIP, GeoRSCLIP, SkySense, GRAFT, SkyEyeGPT, VHM, LHRS}, which often leads to suboptimal performance when handling tasks of different granularities. For example, image-level tasks mainly focus on global features. If local features are overly emphasized, they may interfere with the VLM's global description or attention focus. In contrast, region-level and pixel-level tasks require clear and accurate local details, and complex backgrounds may result in model misjudgments. Moreover, using high-resolution images for image-level tasks, or retaining high resolution in irrelevant areas when performing region-level and pixel-level tasks, can lead to unnecessary computational overhead.

To address these issues, this study proposes a Task-driven Multi-granularity Resolution Adjustment strategy. This strategy dynamically adjusts the input image resolution based on the characteristics of each task, ensuring a balance between accuracy and computational efficiency.

The resolution adjustment process is carried out by the Task Granularity Classifier, as shown in Figure~\ref{fig:2}. In GeoMag, the input query is first encoded by a pre-trained BERT \citep{bert}. The resulting 768-dimensional query representation is then passed through a trained Fully Connected Layers (FCs) module for task classification. The FCs consist of two hidden layers with 256 and 128 neurons, respectively, followed by an output layer with three neurons corresponding to image-level, region-level, and pixel-level task types. This structure not only improves classification accuracy but also ensures computational efficiency.

Once the Task Granularity Classifier produces a prediction, the input image is adjusted accordingly following the strategy described in Equation \ref{eq:2}:
\begin{equation}
\small
  \label{eq:2}
I^{\prime}=
\begin{cases}
I_{100 \times 100}, & G=\text{Image} \\
C_1(I, P) \oplus D(I), & G=\text{Region} \\
C_2\left(C_1(I, P), P\right) \oplus D\left(C_1(I, P)\right) \oplus D(D(I)), & G=\text{Pixel}
\end{cases}
\end{equation}

Here, $G \in \{$ Image, Region, Pixel $\}$ denotes the task granularity. $ I_{100 \times 100}$ represents adjusting the resolution of the input image $I$ to 100×100. $C_n(I, P)$ indicates the n-th instance of Prompt-guided Semantic-aware Cropping (see Section 3.2.2), which uses the prompt $P$ to locate task-relevant regions in the image $I$ and perform cropping, with each successive crop covering a smaller area than the previous one. $\oplus$ refers to image stitching, where the cropped parts of the image are merged back into their corresponding positions in the original image. In addition, the compression function $D(\cdot)$ for an input with height $H$ and width $W$ can be defined as Equation \ref{eq:3}:
\begin{equation}
  \label{eq:3}
D(\cdot)=\operatorname{Resize}\left(\operatorname{Resize}\left(\cdot, \frac{H}{4} \times \frac{W}{4}\right), \mathrm{H} \times W\right)
\end{equation}

By adaptively adjusting the image input, Task-driven Multi-granularity Resolution Adjustment significantly improves GeoMag's efficiency across tasks of different granularities, reduces the demand for computational resources.

\subsubsection{Prompt-guided Semantic-aware Cropping}

Current VLMs exhibit limited capability in perceiving and accurately parsing small visual details \citep{EarthMarker, GeoChat}. To address the challenges RS-VLMs face in detecting small objects and to enhance their ability to focus on specific regions, GeoMag introduces Prompt-guided Semantic-aware Cropping. This strategy employs a gradient-weighted attention map generation method to query key regions in the image, emphasizing the deep semantic correlation between the prompt and the image content.

In implementation, GeoMag obtains the attention distribution over image tokens from the vision-language model's joint representation and weights it with the gradient of the loss function to enhance the response to image regions that most influence the final prediction. During the forward pass of the VLM, by setting "output\_attentions=True", the model outputs attention matrices at each layer, reflecting the relevance between individual image tokens and text tokens.

GeoMag selects the attention weights from the last layer of the model to capture deep semantic alignment between image content and textual tasks. Next, it computes the gradient of the cross-entropy loss with respect to this attention distribution and applies a ReLU activation function to retain positive gradient signals. These activated gradients are then element-wise multiplied with the original attention matrix to obtain a gradient-weighted attention map.

\begin{table*}[h]
\centering
\begin{tabular}{ccccccc}
\hline
\multirow{2}{*}{\textbf{Method}} & \multicolumn{3}{c}{\textbf{RRSIS-D}} & \multicolumn{3}{c}{\textbf{RefSegRS}} \\ \cline{2-7}
~ & \textbf{P@0.5} & \textbf{OIoU} & \textbf{MIoU} & \textbf{P@0.5} & \textbf{OIoU} & \textbf{MIoU} \\ \hline
RRN \citep{RRN} & 51.07 & 66.43 & 45.64 & 31.21 & 66.12 & 43.34 \\ 
CSMA \citep{CSMA} & 55.32 & 69.39 & 48.54 & 28.07 & 64.53 & 41.47 \\ 
LSCM \citep{LSCM} & 56.02 & 69.05 & 49.92 & 32.12 & 63.21 & 38.64 \\
CMPC \citep{CMPC} & 55.83 & 69.22 & 49.24 & 26.57 & 61.25 & 33.57 \\ 
BRINET \citep{BRINET} & 56.90 & 69.88 & 49.65 & 22.56 & 60.16 & 32.87 \\
CMPC+ \citep{CMPC+} & 57.65 & 68.64 & 50.24 & 51.27 & 68.23 & 54.21 \\ 
LGCE \citep{RefSegRS} & 67.65 & 76.34 & 59.37 & \underline{73.75} & \underline{76.81} & \underline{59.96} \\ 
LAVT \citep{LAVT} & 69.52 & 77.19 & 61.04 & 71.44 & 76.46 & 57.74 \\ 
RMSIN \citep{rseg2} & \underline{74.26} & \underline{77.79} & \underline{64.20} & 72.26 & 76.29 & 59.63 \\ \hline
\rowcolor{blue!6}
\textbf{GeoMag} (Ours) & \textbf{81.30} & \textbf{82.67} & \textbf{65.71} & \textbf{77.50} & \textbf{80.27} & \textbf{61.39} \\ \hline
\end{tabular}
\caption{Evaluation results of the Referring Segmentation task. The best results are bolded, and the second-best results are underlined. The evaluation metrics include Precision at an IoU threshold of 0.5 (P@0.5), Overall Intersection over Union (OIoU), and Mean Intersection over Union (MIoU).}
\label{table:1}
\end{table*}

Finally, GeoMag averages the attention scores for each image token region to construct a 2D attention heatmap $H$, which guides the cropping operation. $H$ is formulated as Equation \ref{eq:4}:
\begin{equation}
  \label{eq:4}
  H=\frac{1}{N} \sum_{i=1}^N\left(\operatorname{ReLU}\left(\frac{\partial L}{\partial A_i}\right) \odot A_i\right)
\end{equation}

Here, $N$ denotes the number of image tokens, $L$ is the cross-entropy loss, $A_i$ represents the attention weights of each image token, $\frac{\partial L}{\partial A_i}$ refers to the gradient of the cross-entropy loss with respect to each attention element, and $\odot$ indicates element-wise multiplication.

After obtaining the gradient-weighted attention heatmap, GeoMag draws inspiration from the sliding window approach used in YOLO \citep{YOLO} to generate bounding boxes. Specifically, GeoMag first generates multiple grid cells and creates candidate bounding boxes based on the attention values within each cell, predicting their positions and sizes. Then, by analyzing the total attention within each candidate box and comparing it with the average total attention in neighboring regions, it selects the candidate region with the greatest difference. This process ensures that the final selected bounding box accurately captures the region in the image that is most relevant to the prompt and most likely to contain the target object. Once the final bounding box is obtained, the image is cropped accordingly and used in the stitching operation described in Equation \ref{eq:2}.

By combining gradient weighting with the attention map generation strategy, Prompt-guided Semantic-aware Cropping enhances RS-VLM's ability to focus on key regions and small objects within the image. 

\section{Evaluation}
\subsection{Experimental Setup}
\textbf{Implementation Details:} All experiments related to GeoMag were conducted on a single NVIDIA 80G A100 GPU. The frozen pre-trained models used in GeoMag's pipeline included BERT-base \citep{bert} in the Task Granularity Classifier and SAM2 \citep{sam2} as the Vision Encoder. The base VLM used 7B LLaVA-v1.5 \citep{llava1.5}, and was fine-tuned using LoRA \citep{LoRA}, with the rank set to 8. The Pixel Decoder and Text Projection were both trainable to better adapt to tasks in the remote sensing domain.\\
\textbf{Tasks and Datasets:} For Pixel-level tasks: Referring Segmentation was evaluated on RRSIS-D \citep{rseg2} and RefSegRS \citep{RefSegRS}; Grounded Conversation Generation was evaluated on GeoPixelD \citep{GeoPixel}. For Region-level tasks: Referring Object Classification was evaluated on DIOR-RSVG \citep{DIOR}; Visual Question Answering was evaluated on RSVQA-LR and RSVQA-HR \citep{RSVQA}. For Image-level tasks: Image Caption was evaluated on UCM-Captions \citep{UCM_SYD} and Sydney-Captions \citep{UCM_SYD}; Image Scene was evaluated on AID \citep{AID} and RESISC45 \citep{RESISC45}.

\subsection{Main Results}

\begin{table}[h]
\centering
\resizebox{1\linewidth}{!}{
    \begin{tabular}{cccccc}
    \hline
        \textbf{Method} & \textbf{METEOR} & \textbf{CIDEr} & \textbf{AP50} & \textbf{MIoU} & \textbf{Recall} \\ \hline
        LISA \citep{LISA} & 22.30 & 146.00 & 8.50 & 42.70 & 29.00 \\ 
        PixelLM \citep{PixelLM} & 22.50 & 183.00 & 10.50 & 42.40 & 29.60 \\ 
        GLaMM \citep{GLaMM} & 23.00 & 157.00 & 12.50 & 46.40 & 32.80 \\ 
        GeoPixel \citep{GeoPixel} & \underline{24.00} & \underline{216.00} & \underline{19.00} & \underline{52.30} & \underline{38.80} \\ \hline
        \rowcolor{blue!6}
        \textbf{GeoMag} (Ours) & \textbf{27.62} & \textbf{235.17} & \textbf{20.76} & \textbf{53.96} & \textbf{40.21} \\ \hline
    \end{tabular}
}
\caption{Evaluation results of the Grounded Conversation Generation task. The best results are bolded, and the second-best results are underlined.}
\label{table:2}
\end{table}

\begin{table*}[ht]
\centering
    \begin{tabular}{cccccccccc}
\hline
\multirow{2}{*}{\textbf{Method}} & \multicolumn{3}{c}{\textbf{RSVQA-LR}} & \multicolumn{3}{c}{\textbf{RSVQA-HR TEST1}} & \multicolumn{3}{c}{\textbf{RSVQA-HR TEST2}} \\ \cline{2-10}
~ & \textbf{P} & \textbf{C} & \textbf{A} & \textbf{P} & \textbf{C} & \textbf{A} & \textbf{P} & \textbf{C} & \textbf{A} \\ \hline
\multicolumn{10}{c}{\textbf{Specialized Model (Supervised Evaluation)}} \\ \hline
EasyToHard \citep{EasyToHard} & 90.66 & 87.49 & 89.08 & 91.39 & 89.75 & 90.57 & 87.97 & 87.68 & 87.83 \\
Bi-Modal \citep{Bi-Modal} & 91.06 & 91.16 & 91.11 & 92.03 & 91.83 & 91.93 & 89.37 & 89.62 & 89.50 \\ 
SHRNet \citep{SHRNet} & 91.03 & 90.48 & 90.76 & 92.45 & 91.68 & 92.07 & 89.81 & 89.44 & 90.63 \\ \hline
\multicolumn{10}{c}{\textbf{MLLM (Zero-shot)}} \\ \hline
MiniGPT-4 \citep{MiniGPT-4} & 43.86 & 57.55 & 50.71 & 52.91 & 54.76 & 53.84 & 50.43 & 52.60 & 51.52 \\ 
Shikra \citep{Shikra} & 46.47 & 60.31 & 53.39 & 58.85 & 57.40 & 58.13 & 57.28 & 56.63 & 57.00 \\ 
MiniGPT-v2 \citep{MiniGPT-v2} & 49.85 & 63.09 & 56.47 & 64.80 & 59.17 & 61.98 & 40.79 & 50.91 & 45.85 \\ \hline
\multicolumn{10}{c}{\textbf{RS-VLM (Supervised Evaluation, Gray Indicates Zero-shot)}} \\ \hline
RSGPT \citep{RSGPT} & \underline{91.17} & \underline{91.70} & \textbf{92.29} & \textbf{91.86} & \textbf{92.00} & \textbf{92.00} & \textbf{89.87} & \textbf{89.68} & \textbf{89.78} \\ 
GeoChat \citep{GeoChat} & 91.09 & 90.33 & 90.70 & - & - & - & \cellcolor{gray!20}58.45 & \cellcolor{gray!20}83.19 & \cellcolor{gray!20}72.30 \\ 
EarthGPT \citep{EarthGPT} & - & - & - & - & - & - & \cellcolor{gray!20}62.77 & \cellcolor{gray!20}79.53 & \cellcolor{gray!20}72.06 \\ 
LHRS-Bot \citep{LHRS} & 88.51 & 90.00 & 89.19 & - & - & - & - & - & - \\ 
VHM \citep{VHM} & 90.11 & 89.89 & 89.33 & - & - & - & \cellcolor{gray!20}64.00 & \cellcolor{gray!20}83.50 & \cellcolor{gray!20}73.75 \\
SkyEyeGPT \citep{SkyEyeGPT} & 88.93 & 88.63 & 88.78 & 84.95 & 85.63 & 85.29 & \underline{83.50} & 80.28 & 81.89 \\ \hline
\rowcolor{blue!6}
\textbf{GeoMag} (Ours) & \textbf{91.38} & \textbf{91.81} & \underline{91.59} & \underline{87.04} & \underline{86.81} & \underline{86.92} & 77.62 & \underline{88.94} & \underline{83.27} \\ \hline
\end{tabular}
\caption{Evaluation results of the VQA task. Among RS-VLMs, the best results are bolded, and the second-best results are underlined. The RSVQA-HR test set contains two subsets, which are reported separately. The evaluation metrics include Presence (P),  Comparison (C), and Average Accuracy (A).}
\label{table:3}
\end{table*}

\begin{table}[h]
\centering
\begin{tabular}{ccc}
\hline
\textbf{Method} & \textbf{SS} & \textbf{SIoU} \\ \hline
GeoChat \citep{GeoChat} & 79.59 & 68.80 \\ 
Sphinx \citep{SPHINX} & 93.72 & 89.37 \\ 
EarthGPT \citep{EarthGPT} & \underline{94.64} & \underline{90.16} \\ \hline
\rowcolor{blue!6}
\textbf{GeoMag} (Ours) & \textbf{97.84} & \textbf{96.47} \\ \hline
\end{tabular}
\caption{Evaluation results of the Referring Object Classification task. The best results are bolded, and the second-best results are underlined. The evaluation metrics include Semantic Similarity (SS) and Semantic Intersection over Union (SIoU).}
\label{table:4}
\end{table}

\textbf{Pixel-level tasks.} As shown in Table~\ref{table:1}, GeoMag achieves the best performance among the 10 compared models on the Referring Segmentation task. Compared to the second-best model, P@0.5 improves by 7.04\% and 3.75\% on the two datasets, respectively; OIoU improves by 4.88\% and 3.46\%, respectively; and MIoU improves by 1.51\% and 1.43\%, respectively. These results fully demonstrate GeoMag's advantages in fine-grained semantic understanding, region localization, and pixel-level processing. As shown in Table~\ref{table:2}, GeoMag also achieves the best performance on the Grounded Conversation Generation task. Among the compared models, LISA, PixelLM, and GLaMM are all fine-tuned on the GeoPixelD training set to adapt to remote sensing tasks. Compared to the second-best model, GeoMag improves by 19.17 in CIDEr, the most important metric in the text modality, and by 1.66\% in MIoU, the key metric in the visual modality, indicating GeoMag's stronger modeling capability in image-text alignment and collaborative understanding in the remote sensing domain.

GeoMag not only fills the gap in pixel-level task handling among existing RS-VLMs, but also demonstrates its strengths in recognition and segmentation tasks in the remote sensing field through precise region-focused processing.

\textbf{Region-level tasks.} The evaluation results for the VQA task are shown in Table~\ref{table:3}. The main evaluation metrics for this task include presence (e.g., "Is a large commercial building present in the image?"), comparison (e.g., "Is the number of buildings equal to the number of water areas?"), and Average Accuracy. Among RS-VLMs, GeoMag ranks second in Average Accuracy across all three test sets, with other metrics also ranking between 1st and 3rd, demonstrating strong VQA performance. However, in the VQA task, the performance of RS-VLMs is generally limited by their adaptability to cross-domain tasks and tends to be lower than that of Specialized Models optimized for specific tasks. Therefore, future RS-VLMs should further enhance fine-tuning for specific tasks to narrow the performance gap with specialized models.

The evaluation results for Referring Object Classification are shown in Table~\ref{table:4}. SS measures the semantic similarity between predicted and ground truth labels, while SIoU reflects the degree of overlap between words. GeoMag achieves the best results on the DIOR-RSVG dataset, improving SS by 3.20\% and SIoU by 6.31\% compared to the second-best model. These results fully demonstrate GeoMag's superiority in region-level tasks.

\begin{table*}[h]
\centering

\begin{tabular}{ccccccccc}
\toprule
\textbf{Dataset} & \textbf{Method} & \textbf{BLEU1} & \textbf{BLEU2} & \textbf{BLEU3} & \textbf{BLEU4} & \textbf{METEOR} & \textbf{ROUGE\_L} & \textbf{CIDEr} \\
\midrule

\multirow{13}{*}{UCM} 
& \multicolumn{8}{c}{\textbf{Specialized Model (Supervised Evaluation)}} \\ \cline{2-9}
& SAA \citep{SAA} & 79.62 & 74.01 & 69.09 & 64.77 & 38.59 & 69.42 & 294.51 \\
& SD-RSIC \citep{SD-RSIC} & 74.80 & 66.40 & 59.80 & 53.80 & 39.00 & 69.50 & 213.20 \\
& Post-processing \citep{Post} & 79.73 & 72.98 & 67.44 & 62.62 & 40.80 & 74.06 & 309.64 \\ \cline{2-9}

& \multicolumn{8}{c}{\textbf{MLLM (Zero-shot)}} \\ \cline{2-9}
& MiniGPT-4 \citep{MiniGPT-4} & 30.90 & 27.55 & 22.23 & 18.10 & 33.36 & 41.37 & 0.03 \\
& Shikra \citep{Shikra} & 81.16 & 58.94 & 43.26 & 33.98 & 32.56 & 56.73 & 56.69 \\
& MiniGPT-v2 \citep{MiniGPT-v2} & 81.10 & 60.27 & 45.10 & 36.16 & 32.41 & 56.57 & 60.66 \\ \cline{2-9}

& \multicolumn{8}{c}{\textbf{RS-VLM (Supervised Evaluation)}} \\ \cline{2-9}
& RSGPT \citep{RSGPT} & 86.12 & \underline{79.14} & 72.31 & 65.74 & 42.21 & 78.34 & 333.23 \\
& RS-CapRet \citep{RS-CapRet} & 84.30 & 77.90 & 72.20 & \underline{67.00} & \underline{47.20} & \textbf{81.70} & \textbf{354.80} \\
& SkyEyeGPT \citep{SkyEyeGPT} & \textbf{90.71} & \textbf{85.69} & \textbf{81.56} & \textbf{78.41} & 46.24 & 79.49 & 236.75 \\ \cline{2-9}
\rowcolor{blue!6}
& \textbf{GeoMag} (Ours) & \underline{87.81} & 79.05 & \underline{72.93} & 66.81 & \textbf{47.22} & \underline{80.04} & \underline{346.42} \\

\midrule

\multirow{13}{*}{Sydney} 
& \multicolumn{8}{c}{\textbf{Specialized Model (Supervised Evaluation)}} \\ \cline{2-9}
& SAA \citep{SAA} & 68.82 & 60.73 & 52.94 & 45.39 & 30.49 & 58.20 & 170.52 \\
& SD-RSIC \citep{SD-RSIC} & 72.40 & 62.10 & 53.20 & 45.10 & 34.20 & 63.60 & 139.50 \\
& Post-processing \citep{Post} & 78.37 & 69.85 & 63.22 & 57.17 & 39.49 & 71.06 & 255.53 \\ \cline{2-9}

& \multicolumn{8}{c}{\textbf{MLLM (Zero-shot)}} \\ \cline{2-9}
& MiniGPT-4 \citep{MiniGPT-4} & 29.53 & 25.85 & 20.27 & 16.38 & 32.02 & 42.73 & 0.07 \\
& Shikra \citep{Shikra} & 77.52 & 53.19 & 36.98 & 27.82 & 29.42 & 53.27 & 26.79 \\
& MiniGPT-v2 \citep{MiniGPT-v2} & 77.35 & 55.81 & 40.58 & 32.31 & 29.92 & 52.13 & 33.78 \\ \cline{2-9}

& \multicolumn{8}{c}{\textbf{RS-VLM (Supervised Evaluation)}} \\ \cline{2-9}
& RSGPT \citep{RSGPT} & 82.26 & 75.28 & 68.57 & 62.23 & 41.37 & 74.77 & \textbf{273.08} \\
& RS-CapRet \citep{RS-CapRet} & 78.70 & 70.00 & 62.80 & 56.40 & 38.80 & 70.70 & 239.20 \\
& SkyEyeGPT \citep{SkyEyeGPT} & \textbf{91.85} & \textbf{85.64} & \textbf{80.88} & \textbf{77.40} & \textbf{46.62} & \textbf{77.74} & 181.06 \\ \cline{2-9}
\rowcolor{blue!6}
& \textbf{GeoMag} (Ours) & \underline{85.00} & \underline{78.11} & \underline{71.62} & \underline{64.45} & \underline{43.57} & \underline{75.62} & \underline{259.30} \\

\bottomrule
\end{tabular}
\caption{Evaluation results of the Image Caption task. The best result for each metric is bolded, and the second-best is underlined.}
\label{table:5}
\end{table*}

\begin{table}[h]
\centering
\begin{tabular}{ccc}
\toprule
\textbf{Method} & \makecell{\textbf{AID} \\ \textbf{Accuracy}} & \makecell{\textbf{RESISC-45} \\ \textbf{Accuracy}} \\
\midrule
RemoteCLIP \citep{RemoteCLIP} & \textbf{87.90} & \textbf{79.84} \\
GeoRSCLIP \citep{GeoRSCLIP} & 73.72 & 71.89 \\
GeoChat \citep{GeoChat} & 72.03 & - \\
SkyCLIP \citep{SkyScript} & 71.70 & 70.94 \\
\rowcolor{blue!6}
\textbf{GeoMag} (Ours) & \underline{83.03} & \underline{77.62} \\
\bottomrule
\end{tabular}
\caption{Evaluation results of the Image Scene task. All methods shown are RS-VLMs evaluated under zero-shot settings for this task. The best results are bolded, and the second-best results are underlined.}
\label{table:6}
\end{table}

\textbf{Image-level tasks.} As shown in Table~\ref{table:5}, on the UCM dataset, GeoMag achieves the second-best result among all compared methods on the most critical evaluation metric for the Image Caption task, CIDEr, with other metrics also performing well, ranking between 1st and 3rd. On the Sydney dataset, GeoMag achieves second-best performance across all metrics, further validating its robustness and generalization ability in remote sensing Image Caption tasks. The evaluation results for the Image Scene task are shown in Table~\ref{table:6}. On both the AID and RESISC-45 datasets, GeoMag achieves the second-best zero-shot scene classification accuracy, demonstrating its capability to effectively recognize complex remote sensing scene semantics without the need for additional annotations.
Although GeoMag is primarily designed for more challenging Pixel-level remote sensing image parsing tasks, it still delivers excellent overall performance in Image-level tasks. Tables~\ref{table:5} and ~\ref{table:6} compare 14 different specialized models, general-purpose MLMMs, and existing RS-VLMs, and GeoMag consistently maintains competitive results, reflecting its applicability and transferability in coarse-grained tasks.

\subsection{Ablation Studies}

\begin{table*}[ht]
\centering
\begin{tabular}{c c c c c c c c c c c c}
\toprule
\multirow{2}{*}{\textbf{Method}} & \multicolumn{3}{c}{\textbf{RRSIS-D}} & \multicolumn{3}{c}{\textbf{RefSegRS}} & \multicolumn{2}{c}{\textbf{DIOR-RSVG}} & \multicolumn{3}{c}{\textbf{RSVQA-HR TEST1}} \\
\cmidrule(lr){2-4} \cmidrule(lr){5-7} \cmidrule(lr){8-9} \cmidrule(lr){10-12}
& \textbf{P@0.5} & \textbf{OIoU} & \textbf{MIoU} & \textbf{P@0.5} & \textbf{OIoU} & \textbf{MIoU} & \textbf{SS} & \textbf{SIoU} & \textbf{P} & \textbf{C} & \textbf{A} \\
\midrule
GeoMag w/o MagCrop & 65.45 & 74.20 & 57.47 & 44.93 & 62.74 & 48.51 & 75.04 & 64.29 & 63.97 & 78.93 & 71.45 \\
\rowcolor{blue!6}
GeoMag & 81.30 & 82.67 & 65.71 & 77.50 & 80.27 & 61.39 & 97.84 & 96.47 & 87.04 & 86.81 & 86.92 \\
\bottomrule
\end{tabular}
\caption{Ablation Studies on Pixel-level and Region-level tasks. Abbreviations are the same as in Tables ~\ref{table:1}, ~\ref{table:3}, and ~\ref{table:4}.}
\label{table:7}
\end{table*}

\begin{figure*}[t]
  \includegraphics[width=1\linewidth]{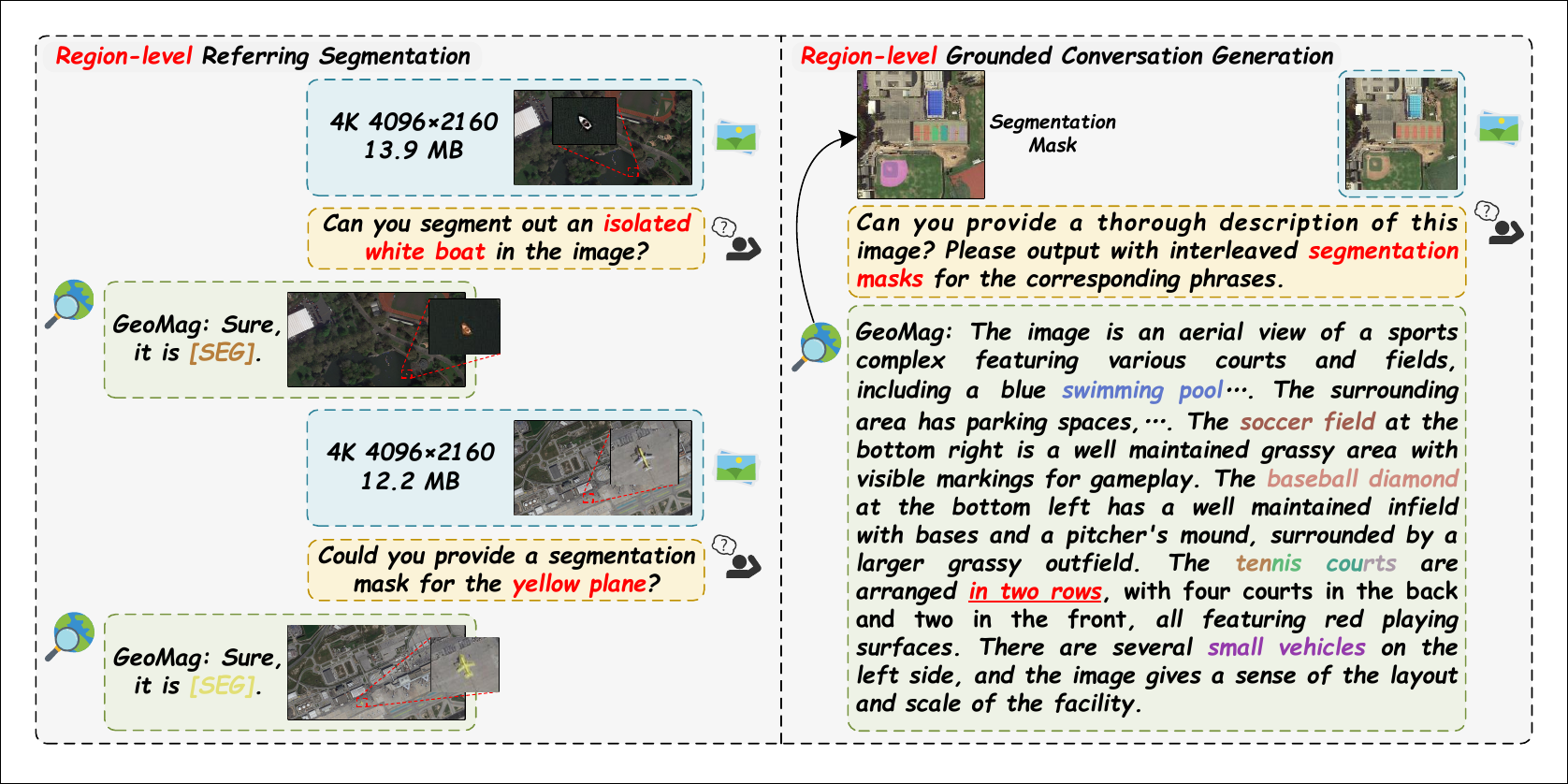}
  \caption {A case study of GeoMag on pixel-level tasks. On the left, GeoMag performs the Referring Segmentation task, accurately segmenting two extremely small targets—a ship and an airplane—in a large-scale 4K input image. On the right, during the Grounded Conversation Generation task, MagCrop provides a detailed description of the input image and outputs the corresponding segmentation masks.}
  \label{fig:4}
\end{figure*}

This study conducted ablation studies on Region-level and Pixel-level tasks to verify whether MagCrop enables RS-VLMs to more effectively address the challenges of diverse task granularities and difficulty in recognizing small objects in remote sensing image analysis. It is worth noting that Grounded Conversation Generation was excluded from ablation testing since it required generating comprehensive general descriptions of the image and did not use MagCrop for preprocessing. As shown in Table~\ref{table:7}, for all three Region-level and Pixel-level tasks, performance drops significantly when MagCrop is not used, which strongly demonstrates its effectiveness. In particular, the tasks Referring Object Classification and Referring Segmentation show severe declines (SIoU drops by 32.18\%, and P@0.5 drops by 15.85\% and 32.57\%, respectively). This is because each question in these tasks typically includes specific object or directional cues, and MagCrop is especially well-suited to such prompts—using prompt-based queries to identify key image regions, focus on target objects, and suppress interference from complex backgrounds. The ablation studies indicate that MagCrop provides significant advantages in enhancing the model's capability for target localization and pixel-level processing.

\subsection{Pixel-level Parsing: A Case Study and Discussion}

Figure~\ref{fig:4} illustrates the performance of GeoMag on pixel-level tasks. In the Referring Segmentation task, GeoMag supports remote sensing image inputs with a 4K resolution (4096×2160), where locating small targets at such a scale is particularly challenging. When prompted with "the isolated white ship" and "the yellow airplane," GeoMag accurately identifies these two small objects and generates the corresponding segmentation masks, demonstrating its strong pixel-level processing capabilities for small target detection. In the Grounded Conversation Generation task, GeoMag produces a detailed description of the input image, covering both the overall scene and representative objects within it, and outputs pixel-level segmentation masks for the corresponding objects. As shown in Figure~\ref{fig:4}, both the textual descriptions and the masks produced by GeoMag are generally accurate. Future research may explore generating more complete segmentation masks. 

However, GeoMag still exhibits certain limitations. As shown in the underlined portion of the textual description on the right side of Figure~\ref{fig:4}, although GeoMag successfully identifies six tennis courts and correctly determines the count, the original image presents the courts arranged in a single row, whereas the generated text describes them as divided into two rows. This discrepancy indicates that GeoMag has room for improvement in handling spatial layouts and relative positional relationships between objects. In the future, enhancing the model's understanding of spatial relationships—such as by incorporating spatial layout modeling mechanisms—can improve the accuracy and coherence of its descriptions. Such improvements will help strengthen GeoMag's consistency in spatial expression and semantic understanding when interpreting complex scenes.

Moreover, there is currently a lack of available high-quality datasets for pixel-level tasks in remote sensing image analysis. This issue is particularly evident in the Grounded Conversation Generation task, which remains in an early exploratory stage. Therefore, creating and releasing more high-quality datasets will be essential for advancing the development of pixel-level tasks and should become a key focus for future research efforts.

\section{Conclusion}
This study proposes GeoMag, a RS-VLM capable of handling multi-granularity RS image parsing tasks. By introducing components such as a Pixel Decoder, GeoMag extends its capabilities to pixel-level processing. It further presents MagCrop, an image preprocessing strategy that combines Task-driven Multi-granularity Resolution Adjustment with Prompt-guided Semantic-aware Cropping. This strategy allows the model to automatically perceive the task type and apply the appropriate processing method, thereby enhancing the accuracy of small object recognition and pixel-level segmentation, while simultaneously reducing computational redundancy by downsampling non-essential regions. Experimental results demonstrate that GeoMag not only achieves effective pixel-level processing but also delivers competitive performance across tasks of other granularities. These contributions provide a novel perspective for advancing remote sensing multimodal understanding toward a unified, all-granularity paradigm.

\begin{acks}
This work was supported by the National Natural Science Foundation of China (Grant No.W2412136), in part by the Postgraduate Research \& Practice Innovation Program of Jiangsu Province (Grant No.KYCX25\_0343), in part by the National Key Research and Development Program of China (Grant No.2022YFB2901800), in part by the National Natural Science Foundation of China (Grant No.62202445), in part by the National Natural Science Foundation of China-Research Grants Council (RGC) Joint Research Scheme (Grant No.62321166652), in part by the Fundamental Research Funds for the Central Universities (Grant No.491914380007), and in part by the 'GeoX' Interdisciplinary Project of Frontiers Science Center for Critical Earth Material Cycling (Grant No.20250106).
\end{acks}

\bibliographystyle{ACM-Reference-Format}
\balance
\bibliography{sample-base}
\end{document}